\documentclass[10pt,twocolumn,letterpaper]{article}

\usepackage{iccv}
\usepackage{times}
\usepackage{epsfig}
\usepackage{graphicx}
\usepackage{amsmath}
\usepackage{amssymb}
\usepackage{booktabs}
\usepackage{multirow}
\usepackage{bm}
\usepackage{algorithm}
\usepackage{algpseudocode}
\usepackage{color}
\usepackage[breaklinks=true,bookmarks=false]{hyperref}

\iccvfinalcopy % *** Uncomment this line for the final submission

\def\iccvPaperID{****} % *** Enter the ICCV Paper ID here

\ificcvfinal\fi

\begin{document}

%%%%%%%%% TITLE
\title{Hierarchical Representation based Query-Specific Prototypical Network for Few-Shot Image Classification}

\author{Yaohui Li, Huaxiong Li, Haoxing Chen, Chunlin Chen\\
Nanjing University,
Nanjing, China\\
{\tt\small \{yaohuili, haoxingchen\}@smail.nju.edu.cn, \{huaxiongli, clchen\}@nju.edu.cn}
% For a paper whose authors are all at the same institution,
% omit the following lines up until the closing ``}''.
% Additional authors and addresses can be added with ``\and'',
% just like the second author.
% To save space, use either the email address or home page, not both

}
\maketitle
% Remove page # from the first page of camera-ready.
\ificcvfinal\thispagestyle{empty}\fi

%%%%%%%%% ABSTRACT
\begin{abstract}
Few-shot image classification aims at recognizing unseen categories with a small number of labeled training data. Recent metric-based frameworks tend to represent a support class by a fixed prototype (e.g., the mean of the support category) and make classification according to the similarities between query instances and support prototypes. However, discriminative dominant regions may locate uncertain areas of images and have various scales, which leads to the misaligned metric. Besides, a fixed prototype for one support category cannot fit for all query instances to accurately reflect their distances with this category, which lowers the efficiency of metric. Therefore, query-specific dominant regions in support samples should be extracted for a high-quality metric. To address these problems, we propose a Hierarchical Representation based Query-Specific Prototypical Network (QPN) to tackle the limitations by generating a region-level prototype for each query sample, which achieves both positional and dimensional semantic alignment simultaneously. Extensive experiments conducted on five benchmark datasets (including three fine-grained datasets) show that our proposed method outperforms the current state-of-the-art methods.
\end{abstract}

%%%%%%%%% BODY TEXT
\section{Introduction}
Deep neural networks have achieved great success in visual tasks during the past decade\cite{2012ImageNet, 2015ImageNet}. However, its data-driven nature usually suffer from insufficient labeled training data. Besides, in real-world applications, collecting sufficient data and labeling with high confidence become notably time-consuming and expensive. Therefore, many researchers are committed to developing powerful models to learn novel unseen concepts (query samples) from scarce labeled training data (support samples), which is so-called Few-shot Learning (FSL)\cite{fe2003bayesian}. Few-shot image classification is one of the key research directions in the field of FSL, which aims at performing classification on novel categories with only few training data. \par

To address this challenging problem, many meta-learning based approaches have been proposed, which can be broadly divided into two branches, i.e., optimization-based methods\cite{baik2020learning,Santoro2016One, sun2019meta} and metric-based methods\cite{koch2015siamese,
simon2020adaptive, 2020Feat,Zhang2020DeepEMD}. Specifically, optimization-based methods aim to converge the model to novel tasks with few optimization steps, while metric-based approaches tend to learn a transferable deep embedding space and make classification according to the similarities between samples.

%figure1
\begin{figure}[t]
\centering
\includegraphics[width= 0.42\textwidth]{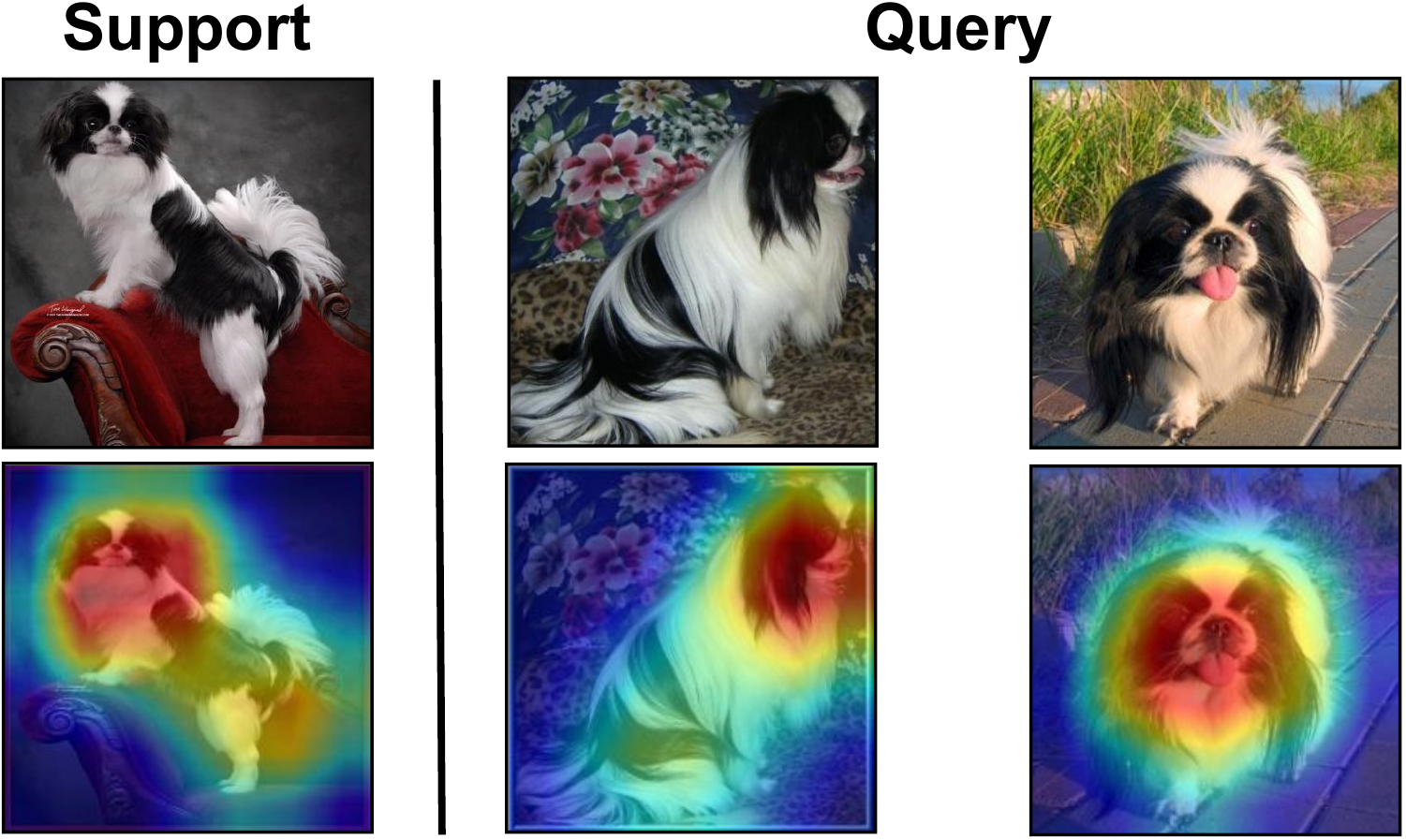}\\
\caption{The images above belong to the same category. Obviously, the position and dimension of the discriminative key regions (red parts) with the same semantic information vary from image to image. Therefore, semantic alignment is necessary before metric and each query sample needs a unique and meaningful measurement with the support category. Our proposed QPN considers about the peculiarity of each query sample by generating query-specific prototypes and achieves both positional and dimensional semantic alignment.} 
\label{fig1}
\end{figure}
Although metric-based approaches have achieved great success on few-shot image classification, existing methods still have several limitations as presented in \cite{0Temperature,hao2019collect}. \par
First, most of the metric-based approaches\cite{hao2019collect, li2019distribution,snell2017prototypical, sung2018learning} utilize a fixed prototype to represent a support category. However, each query sample has its unique distribution of region features, the distance between the query sample and the fixed prototype cannot really show the similarity between them. While the dominant regions of support samples that are close to the regions of the query sample are more likely to contain relevant information with regarding to this query sample\cite{li2019revisiting} and should be extracted to form a query-specific prototype for a more real metric. Therefore, a desirable metric-based method should be able to distinguish dominant regions and extract relevant key regions for each query sample from support categories. For instance, TemperatureNet\cite{0Temperature} re-weights the support samples according to their similarities to the query sample to generate query-specific prototypes, which captures the query-specific information and achieve better performance but the generated instance-level prototype ignores the deep semantic relevance between regions. To address this issue, we propose a novel Query-Specific Prototypical Network (QPN) to generate a region-level prototype for each query sample by taking advantage of the local similarity measurement between local regions. We first utilize the Convolutional Block Attention Module (CBAM)\cite{wooplk18} to capture discriminative key regions, then select relevant regions from the support category for the query regions and re-weight these support regions to form a query-specific prototype.\par
Second, most metric-based approaches conduct straightforward measurement\cite{snell2017prototypical, sung2018learning, zhang2018deep} between query samples and prototypes for classification, regardless of the semantic relevance between local regions. Actually, discriminative key regions for classification can locate anywhere on feature maps\cite{hao2019collect, Zhang2020DeepEMD} (refer to Figure \ref{fig1}). As a result, dominant regions in a query sample are probably compared with semantically irrelevant regions of the support samples, which may cause serious ambiguity to classification accuracy. To address this issue, for instance, SAML\cite{hao2019collect} adopts a relation matrix to collect and select semantically relevant region pairs, which captures the semantic relevance. However, SAML does not remove the influence of irrelevant information and utilizes the fixed class-mean prototype, which lower the efficiency of the metric. Our method only selects query-relevant regions from support samples for semantic alignment and completely eliminates the influence of noise, ensuring the efficiency of metric. \par
\begin{figure*}[t]
\centering
\includegraphics[width= 0.98\textwidth]{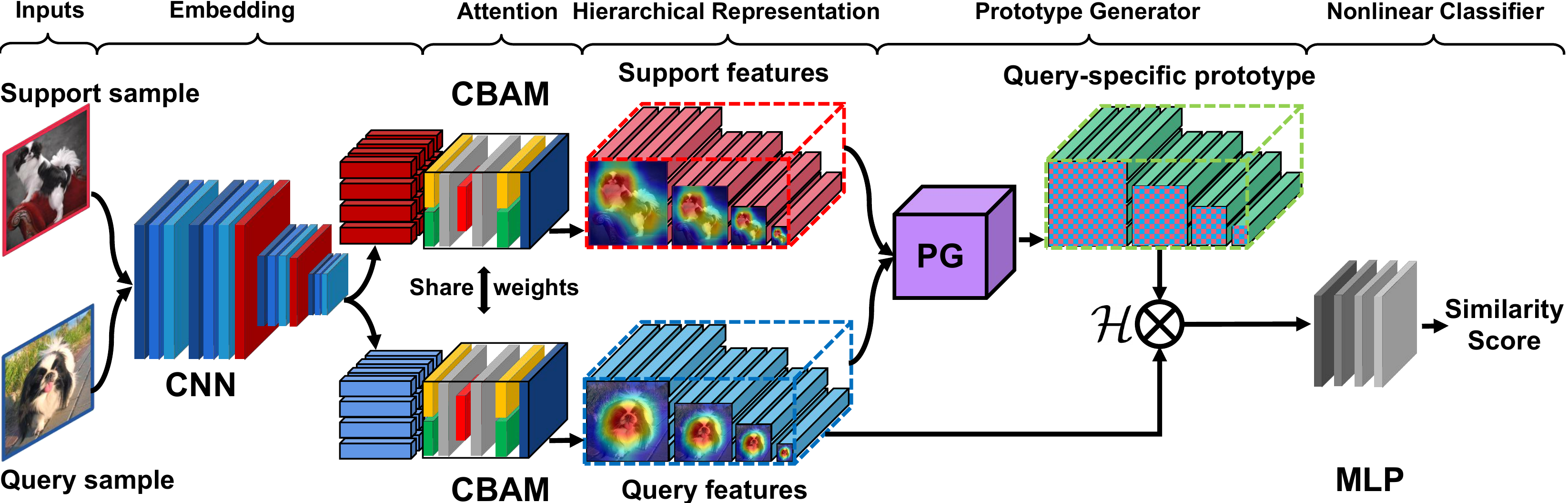}\\
\caption{The overall framework of the proposed QPN under 5-way 1-shot setting.} %The model contains four main parts: the embedding network $\mathcal{F}_{\Theta}$ to obtain feature maps of input images, the \emph{Feature Adjustment Module} consists of an \emph{Attention Block} and a \emph{Dimension Correction Module}, the \emph{Corresponding Prototypical Representation Module} to find pixel-level consistency between query instance and support class, and the non-linear classifier to obtain relation score.}
\label{fig2}
\end{figure*}

Third, almost all FSL methods utilize the single-scale representation\cite{finn2017model-agnostic,li2019distribution, snell2017prototypical, sung2018learning, 2020Feat} to characterize image features. However, dominant objects may have different dimensions in different images (see in Figure \ref{fig1}). As a result, semantic alignment fails when the scale of the dominant object in the query feature map differs greatly from that of the support feature map\cite{hao2019collect}. Inception Operator\cite{2015Going} is an effective solution to this issue, which utilizes additional convolutional layers after embedding network to obtain more comprehensive features with multiple scales. While additional layers introduce more parameters thus make the model much more complex. In this paper, we replace the convolutional layers by simple pooling layers. Specifically, we conduct multi-scale downsampling to the original feature map and get the hierarchical representation with richer scale-wise information (see in Figure \ref{fig2}). Our hierarchical representation mechanism outperforms the single-scale representation and achieves competitive performance with the Inception Operator (see in Table \ref{tab4}).\par
The main contributions of this paper are presented as follows:\par
\begin{itemize}
\item We propose a region-level Prototype Generator (PG) which can generate a semantically aligned high-quality prototype for each query sample.\par
\item We adopt a channel-wise and spatial-wise attention mechanism to capture dominant regions and compress useless regions.\par
\item We propose a hierarchical representation mechanism to obtain more comprehensive features and solve dimensional mismatching. \par
\end{itemize}
\section{Related Works}
%figure2

Few-shot image classification aims at distinguishing novel categories with only a few training samples\cite{fe2003bayesian, fei2006one}. To address the challenging task, many optimization-based and metric-based approaches have been proposed.\par
\subsection{Optimization-based Approaches} 
Optimization-based approaches aim at quickly adapting the model to new tasks with scarce training data. The most representative method of optimization-based methods is MAML\cite{finn2017model-agnostic}, which aims to train a meta learner to find the optimal initialization parameters for classifiers so that the classifier can quickly adapts to new tasks with only a few instances.\par
However, optimization-based approaches usually need costly high-order gradient, which leads to training failure when facing deeper networks\cite{mishra2017simple}. To the contrary, our proposed QPN is much easier for training.
%Meta-learning based approaches train a meta-learner with learning-to-learn paradigm to guide the optimization steps of classifier. Specifically, MANN\cite{Santoro2016One} regards an LSTM as meta-learner to interact with an extra memory module, which both learns optimization strategy and episode-wise initialization for classifier. MAML\cite{finn2017model-agnostic} and its improved version Reptile\cite{nichol2018on} train a meta-learner to find suitable initialization for classifier so that the classifier can quickly adapts to new tasks with only a few instances. MetaOpt\cite{LeeMRS19} adopts the linear classification rule by exploiting the convex optimization problem. MM-Net\cite{cai2018memory} uses memory slots to construct a contextual learner, which predicts the parameters of an embedding network for unlabeled images. MAML+L2F\cite{baik2020learning} proposes a novel initialization concept by employing task-dependent attenuation, which controls the exploitation rate of previous knowledge dynamically.\par 
%However, meta-learning based approaches usually need costly high-order gradient, which leads to training failure when facing deeper networks as is illustrated in \cite{mishra2017simple}. To the contrary, the proposed CPR-Net is trained in an end-to-end manner without introducing external memory module or additional fine-tuning stage.\par
\subsection{Metric-based Approaches}
Our proposed method is most similar to metric-based approaches\cite{koch2015siamese,vinyals2016matching}. Metric-based approaches aim to learn a transferable embedding space, in which the samples from the same category are close to each other and the samples from different categories are relatively far away from each other. \par 
\textbf{Methods based on Fixed Prototypes.} ProtoNet\cite{snell2017prototypical} takes the center point of a support class as its prototype and conducts classification by comparing the distances between the query sample and the prototypes. RelationNet\cite{sung2018learning} aims at learning a metric, which introduces a nonlinear classifier to measure the similarities between prototypes and query instances. CovaMNet\cite{li2019distribution} exploits the more consistent and transferable low-level information by local representations, which takes the second-order covariance representations as prototypes and conducts classification based on the distribution consistency. SAML\cite{hao2019collect} utilizes relation matrix to realize effective feature alignment between query samples and prototypes. FEAT\cite{2020Feat} takes advantage of the set-to-set function (Transformer\cite{vaswani2017attention}) to generate task-specific and discrimnative embeddings.\par
%However, the methods above mainly adopt fixed prototypes without considering the deviations between query instances. Moreover, these methods tend to conduct measurements directly without exploiting the low-level semantic consistency between query instances and support prototypes.\par
%The proposed CPR-Net follows the metric-learning paradigm while makes some improvements. We construct unique CPR for each query instance by searching out the pixel-level semantic consistency between each query instance and support classes, which tackles the two problems above simultaneously. Additionally, we adopt attention mechanism to filter out useless information and dimensional correction to obtain multi-scale LRVs, which both contribute to the construction of better CPRs.\par
\textbf{Methods based on Dynamic Prototypes.} Categorizing query samples by matching them with the closest prototype\cite{snell2017prototypical} is a common practice in few-shot image classification. However, a fixed prototype for one support class is always biased due to the lack of data under few-shot settings, which cannot truly characterize class information and is incompetent for the further metric with various query samples. IMP\cite{AllenSST19} utilizes a set of clusters to represent a support class and the number of clusters is affected by samples. BD-CSPN\cite{LiuSQ20} conducts prototype rectification to find the optimal prototype which have the maximum similarity to all data points within the same class. TemperatureNet\cite{0Temperature} re-weights support samples by the temperature function according to their distances to the query sample and generates query-specific prototypes.\par
Similar to the methods above, our method also generates query-specific prototypes by distance-based selection. While our model selects semantically relevant regions for each local region of a query sample and generates fine-grained prototypes instead of instance-level prototypes. As a result, our model pays more attention to the fine-grained information and thus performs well on fine-grained datasets (see in Table \ref{tabfg}).\par
\subsection{Other FSL Methods}
Besides the two main streams, many other effective methods are proposed to solve few-shot image classification, e.g., methods based on the graph neural networks\cite{garcia2017few-shot, kim2019edge-labeling, liu2019learning}, the reinforcement learning\cite{ChuLCW19} and the convex optimization\cite{LeeMRS19}, etc.
\section{The Proposed Method}
\subsection{Problem Formulation}
%Following the general settings of few-shot image classification, we utilize the widely adopted episodic training mechanism\cite{vinyals2016matching}.\par
%Specifically, few-shot learning problems usually contain three disjoint datasets (e.g. training set, validation set and test set) and each set contains sufficient classes and instances. For each episode, we randomly sample a support set $\mathcal S$ and a query set $\mathcal Q$ from the training/validation/test set. Support set $\mathcal S$ contains $N$ categories with $K$ ($K$ is usually set to 1 or 5) instances in each category. Note that $\mathcal S$ and $\mathcal Q$ have no intersection while share the same label space. Therefore, each episode can be viewed as an independent $N$-way $K$-shot task, in which we want to categorize a query instance $q^i \in \mathcal{Q}$ into $N$ support categories.\par
We start by the fundamental strategies of the standard FSL. In FSL, the so-called episodic training mechanism\cite{vinyals2016matching} is utilized to obtain a well-trained model by sampling a few samples for every episode (task) under the meta-learning paradigm. For each episode, we randomly sample a support set $\mathcal S = \{ (x^{1,1}, l^{1,1}), (x^{1,2},l^{1,2}), \cdots (x^{N,K}, l^{N,K})\}$ and a query set $\mathcal Q= \{ q^1, q^2, \cdots, q^{NM}\}$. Note that $\mathcal S$ and $\mathcal Q$ have no intersection while share the same label space. Therefore, each episode can be viewed as an independent $N$-way $K$-shot task, in which we want to categorize a query instance $q^i \in \mathcal{Q}$ into $N$ support categories.\par
During the training stage, thousands of episodes (tasks) are fed into the model for parameter update and learning transferable knowledge, aiming at generalizing to novel tasks.
\subsection{Overall Framework}
The overall architecture (under 1-shot setting) of our model is shown in Figure \ref{fig2}.\par
Specifically, we first feed a support sample and a query sample into the embedding network and obtain two feature maps. Afterwards, we put the two feature maps through the CBAM to highlight the key regions. Then, we utilize the downsampling operation to rescale the feature maps and finally represent them with the hierarchical representations, which relieves the dimensional mismatching between dominant objects. \par
\begin{figure}[t]
\centering
\includegraphics[width= 0.48\textwidth]{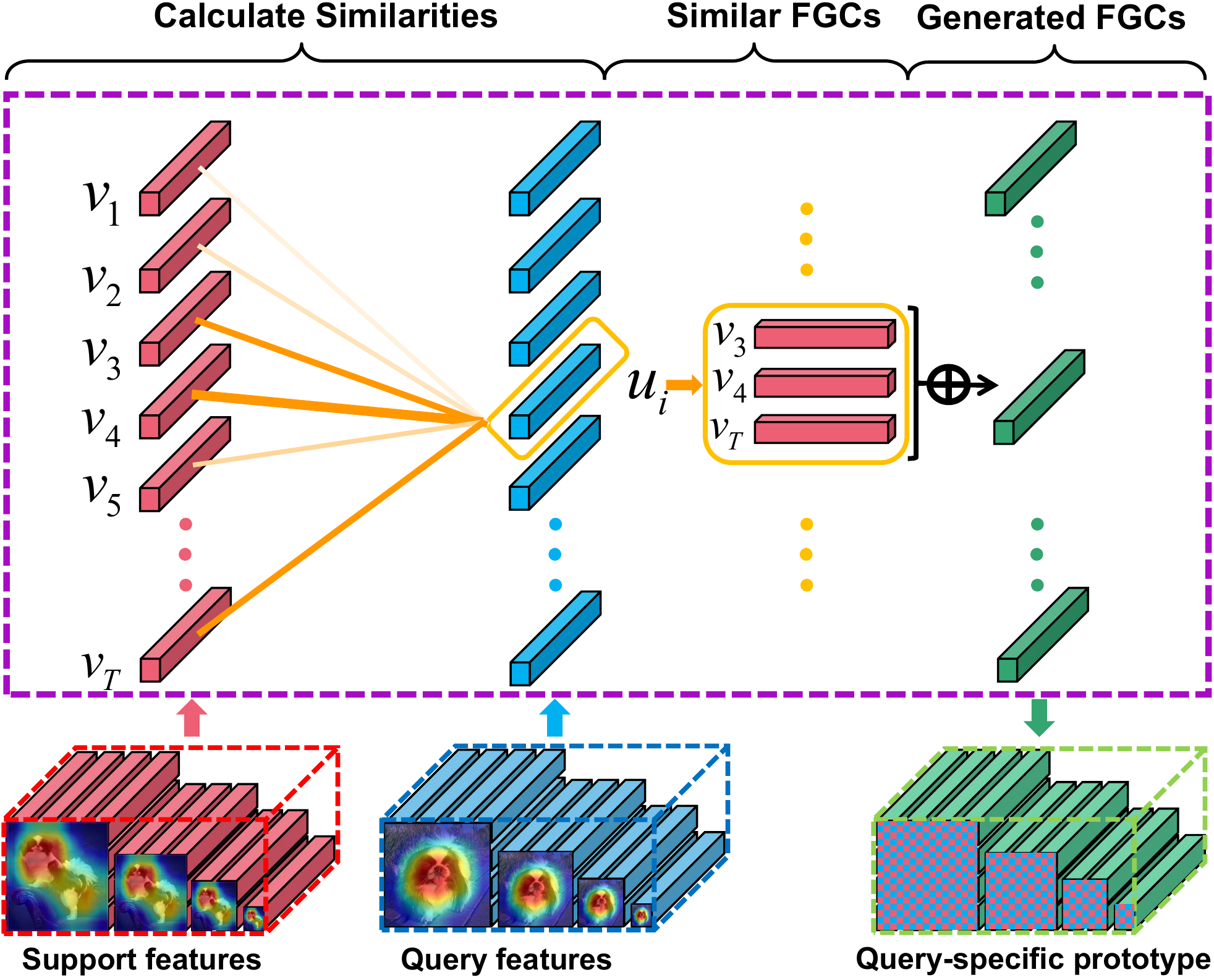}\\
\caption{The detailed architecture of the PG module with ${\xi = 3}$. Note that $\oplus$ denotes the weighted summation.}
\label{fig3}
\end{figure}

Then we put the two hierarchical feature maps into the Prototype Generator (PG), which is the key part of our model (detailed illustration in Figure \ref{fig3}). Specifically, in this module, we find the nearest neighbors\cite{BoimanSI08} of a query region from the support regions and save the weighted sum of the neighbors as a local region of the query-specific prototype. After repeating the above operation to all query regions, a query-specific prototype is completely generated. Finally, we utilize a relation function $\mathcal{H}(,)$ to calculate the relation map and feed it into a nonlinear classifier to obtain the similarity score.\par
%Specifically, we feed one support class $\bm s = \{ x\}$ and one query sample $\bm q$ into the embedding network ${\mathcal{F}}_{\Theta}$ to obtain $\bar{\bm s}$ and $\bar{\bm q}$. Afterwards, we adopt Attention Mechanism ${\mathcal{A}}_{\Omega}$ to highlight discriminative dominant regions and compress semantically irrelevant regions. After attention, $\bar{\bm s}$ and $\bar{\bm q}$ are refined to $\vec{\bm s}$ and $\vec{\bm q}$. For $\vec{\bm s}$ and $\vec{\bm q}$, we correct them by multi-scale pooling and combine all LRVs after correction as $\tilde{\bm s}$ and $\tilde{\bm q}$. In this way, the deviation caused by the dimensional discrepancy of key regions between query and support instances is narrowed to some extent. \par
%In CPR Module, for each LRV of query instance $\tilde{\bm q}$, a similarity based selection is conducted above all LRVs of $\tilde{\bm s}$ (see in Figure \ref{fig3}). By choosing several most similar LRVs from the support class and preserving their weighted summation as corresponding LRV for each LRV in $\tilde{\bm q}$, a pixel-level CPR $\hat{\bm q}$ of the support class $\tilde{\bm s}$ is constructed for the query instance $\tilde{\bm q}$. Obviously, $\hat{\bm q}$ and $\tilde{\bm q}$ are semantically aligned.\par
%Finally, we feed the relation map of $\hat{\bm q}$ and $\tilde{\bm q}$ to a non-linear classifier ${\mathcal{C}}_{\Phi}$ with learnable parameter $\Phi$ to obtain the similarity score.
\subsection{Feature Embedding}
In this work, following the effective methods\cite{snell2017prototypical, sung2018learning, li2019revisiting,2020Multi}, we adopt a four-layer Convolutional Neural Network (CNN) as our embedding network to map all images to a shared representation space. For instance, given an input image $\bm x$, the output of the embedding network is a $3D$ tensor consists of $h\times w$ $c$-dimension vectors, which can be viewed as $h\times w$ fine-grained cells (FGCs). Note that each FGC contains the semantic information of the corresponding local region.\par
\begin{equation}
\bar{\bm x} = {\mathcal{F}}_{\Theta}(\bm x)\in\mathbb{R}^{c\times h \times w},
\end{equation}
where ${\mathcal{F}}_{\Theta}$ is the learned hypothesis function, $\Theta$ denotes the learnable parameters of the neural network, $c$, $h$ and $w$ are the lengths of the three dimensions of the 3D tensor. Concretely, we obtain $c=64$ and $h = w = 21$ in our implementation (441 64-$d$ FGCs) with input size of $84\times 84$. %In addition, following \cite{2017Few, 2019Meta}, we also adopt pre-trained Resnet\cite{he2016deep} for embedding to make more comprehensive comparisons with the state-of-the-arts.%Namely, an input image can be represented as $h \times w$ $c$-dimensional local region vectors(LRVs):
%\begin{equation}
%f^{\bm X} ={\mathcal{F}}_{\Theta}(\bm X)=[{\bm x}_1, {\bm x}_2, \dots, {\bm x}_{hw}]\in \mathbb{R}^{(h\times w)\times c},
%\end{equation}
%where $x_i$ denotes the $i$-$th$ local region. Specifically, given a support class $S$ with $K$ instances and a query instance $q$, we can acquire:
%\begin{equation}
%\begin{aligned}
%&f^{\bm S} = {\mathcal{F}}_{\Theta}(\bm S)= [{\bm s}_1, {\bm s}_2, \dots, {\bm s}_{Khw}],\\&f^{\bm q}= {\mathcal{F}}_{\Theta}({\bm q})= [{\bm q}_1, {\bm q}_2, \c, {\bm q}_{hw}].
%\end{aligned}
%\end{equation}

%figure3
\subsection{Attention} 
Following the classic framework of the CBAM\cite{wooplk18, zhu2020multi}, we conduct both Channel-wise Attention (CWA) and Spatial-wise Attention (SWA) on the input feature maps. Note that the CWA and the SWA are connected in series. Specifically, given an intermediate feature map $\bar{\bm x}\in\mathbb{R}^{c\times h\times w}$ extracted by the embedding network, the CWA attempts to exploit the weight relation between channels. The CWA first conducts spatial squeeze by global average pooling and global max pooling and generates two weight tensors $\bar{\bm x}_{max}^{c}\in\mathbb{R}^{c\times 1\times 1}$ and $\bar{\bm x}_{avg}^{c}\in\mathbb{R}^{c\times 1\times 1}$. Then $\bar{\bm x}_{max}^{c}$ and $\bar{\bm x}_{avg}^{c}$ are combined to generate the channel-wise attention weight $W_{c}$ through a Multilayer Perceptron (MLP):\par
\begin{equation}
W_{c} = \delta \{[MLP_{\omega_c}(\bar{\bm x}_{max}^{c})]+[MLP_{\omega_c}(\bar{\bm x}_{avg}^{c})]\},
\end{equation}
where $\omega_c$ denotes the parameters of the MLP and $\delta$ denotes the sigmoid function. Therefore, the channel-wise attention feature map ${\bar{\bm x}}' = W_c \otimes \bar{\bm x}$, where $\otimes$ denotes the element-wise multiplication and the channel-wise attention weight $W_{c}$ is expanded along the spatial dimension during the multiplication.\par
Afterwards, the SWA exploits the weight relation between spatial regions. In detail, the SWA utilizes the channel squeeze by max pooling and average pooling to obtain two weight maps $ \bar{\bm x}_{max}^{s}\in\mathbb{R}^{1\times h\times w}$, $ \bar{\bm x}_{avg}^{s}\in\mathbb{R}^{1\times h\times w}$. Then $\bar{\bm x}_{max}^{s}$ and $\bar{\bm x}_{avg}^{s}$ are fused through the one-layer convolution to get the spatial-wise attention weight $W_{s}$: \par
\begin{equation}
W_{s} = \delta \{ \mathbb{C}_{\omega_s}[Cat(\bar{\bm x}_{max}^{s}, \bar{\bm x}_{avg}^{s})]\},
\end{equation}
where $\omega_s$ represents the parameters of the $1\times 1$ convolutional layer $\mathbb{C}$ and $Cat(,)$ denotes the operation of the concatenation. The spatial-wise attended feature map can be refined as ${\vec{\bm x}}= W_s \otimes {\bar{\bm x}}'$. During the multiplication, the spatial-wise attention weight $W_{s}$ is expanded along the channel dimension.\par
Through the Attention operation, the key regions of feature maps are highlighted effectively, which is validated in the ablation experiments (see in Table \ref{tab3}).\par 
\subsection{Hierarchical Representation}
As is shown in Figure \ref{fig1}, semantic alignment fails when discriminative key regions are vastly different in dimension\cite{hao2019collect}. Therefore, following the parametric method Inception\cite{ szegedy2015going,hao2019collect}, we put forward a non-parametric hierarchical representation strategy to relieve the dimensional mismatching. Given a feature map after Attention $\vec{\bm x}\in \mathbb{R}^{ c \times h \times w}$, we downsample it to fit the dimensions of the dominant objects in other samples as much as possible. \par
Specifically, based on experimental comparisons and the consideration of the complexity, we rescale the original $21\times 21$ feature map to $10\times10$, $7\times7$, $5\times 5$, $3\times 3$, $2\times 2$ and $1\times 1$, respectively. Namely, original 441 FGCs are expanded to 629 FGCs ($629 =\sum_{i = 10,7,5,3,2,1} i^{2}$ ). Note that our hierarchical representation strategy achieves competitive performance compared with the Inception operator (see in Table \ref{tab4}).\par%Specifically, we adopt average pooling to shrink the spatial dimension:\par
%\begin{equation}
%\begin{aligned}
%&\vec{\bm x}^{(d)} = AP_{d}(\vec{\bm x})\in \mathbb{R}^{ c \times d \times d},\\&d \in \{1, 2, 3, 5, 7, 10\},
%\end{aligned}
%\end{equation}
%where $AP_d$ denotes the average pooling with output size of $d \times d$ in spatial dimension. %In our work, d = \{10, 7, 5, 3, 2, 1\}. 

\subsection{Prototype Generator}

As mentioned before, only one fixed prototype for one support category cannot fit for all query instances. Therefore, different from previous fixed prototypes, we propose a novel Prototype Generator to generate region-level query-specific prototypes of support categories for each query sample. Particularly, given a support category $\bm s$ with $K$ instances and a query instance $\bm q$, their hierarchical representations are as follows:\par 
%\begin{equation}
%\begin{aligned}
%&\vec{\bm s} = {\mathcal{A}_{\Omega}}[{\mathcal{F}}_{\Theta}(\bm s)]\in \mathbb{R}^{K\times c \times h\times w},\\& \vec{\bm q} = {\mathcal{A}_{\Omega}}[{\mathcal{F}}_{\Theta}(\bm q)]\in \mathbb{R}^{ c \times h \times w}.
%\end{aligned}
%\end{equation}\par
%To narrow the dimensional mismatching, we make dimensional correction on the $K$ instances of $\vec{\bm s}$ and the query instance $\vec{\bm q}$. For the convenience of explanations, we represent the corrected $\tilde{\bm s}$ and $\tilde{\bm q}$ with $c$-dimension LRVs\cite{hao2019collect, li2019revisiting}:\par%: $\tilde{\bm s} =[\vec{\bm s}, \vec{\bm s}^{10}, \vec{\bm s}^{7}, \vec{\bm s}^{5},\vec{\bm s}^{3}, \vec{\bm s}^{2},\vec{\bm s}^{1}]$, $\tilde{\bm q} =[\vec{\bm q}, \vec{\bm q}^{10}, \vec{\bm q}^{7}, \vec{\bm q}^{5},\vec{\bm q}^{3}, \vec{\bm q}^{2},\vec{\bm q}^{1}]$. 

\begin{equation}
\begin{aligned}
&\tilde{\bm s} = [{\bm v}_1, {\bm v}_2, \cdots, {\bm v}_{KT}]\in \mathbb{R}^{c\times KT}, \\&\tilde{\bm q} =[{{\bm u}}_1, {{\bm u}}_2, \cdots, {{\bm u}}_{T}]\in\mathbb{R}^{c\times T},
\end{aligned}
\end{equation}
where $T$ denotes the number of FGCs in the hierarchical representation of an instance ($T=629$). Then for each FGC $\bm{u}_i$ of the query instance $\tilde{\bm q} $, we calculate its similarities with all FGCs in $\tilde{\bm s}$ by the similarity function $g$: \par
\begin{equation}
%\{\hat{\bm u}_i^{m}|_{m=1}^{\xi}\} \gets 
\mathcal{G}_i = [g(\bm{u}_i,{\bm v}_1), \cdots, g(\bm{u}_i,{\bm v}_{KT})]. 
\end{equation}
Then we find the $\xi$ maximum similarities in $\mathcal{G}_i$ and record the corresponding ${\xi}$ FGCs from $\tilde{\bm s}$ as: $\{\hat{\bm v}_i^{m}|_{m=1}^{\xi}\}$. Note that $\xi$ is called the generation coefficient and $\{\hat{\bm v}_i^{m}|_{m=1}^{\xi}\}$ are semantically related to $\bm{u}_i$. Then we calculate the weighted sum of $[\hat{\bm v}_i^{1}, \hat{\bm v}_i^{2}, \cdots, \hat{\bm v}_i^{\xi}]$:\par
\begin{equation}
\begin{aligned}
&\hat{\bm v}_i = \sum_{m=1}^{\xi} \bm{\rho}_m \hat{\bm v}_i^{m},\\& \bm{\rho}_m = \frac{exp[g(\bm{u}_i,\hat{\bm v}_i^{m})]}{\sum_{m=1}^{\xi} exp[g(\bm{u}_i,\hat{\bm v}_i^{m})]},\\%&cos(\bm{u}_i,\hat{\bm u}_i^{m}) = \frac{{\bm{u}_i}^{\top} \hat{\bm u}_i^{m}}{||\bm{u}_i||\cdot ||\hat{\bm u}_i^{m}||},
\end{aligned}
\end{equation}
where $\hat{\bm v}_i$ represents the generated FGC corresponding to $\bm{u}_i$, $\bm{\rho}_m$ is the weight of $\hat{\bm v}_i^{m}$ and $g$ denotes the similarity function. After generating a corresponding FGC for each FGC of the query instance $\tilde{\bm q}$, a query-specific prototype (QP) $\hat{\bm q} = [\hat{\bm v}_1, \hat{\bm v}_2, \cdots, \hat{\bm v}_{T}]$ for $\tilde{\bm q}$ is completely generated. Obviously, $\hat{\bm q}$ and $\tilde{\bm q}$ are semantically aligned.\par
It is worth mentioning that the region-level prototypical generation is more powerful than the instance-level prototypical generation\cite{LiuSQ20,0Temperature} in terms of discovering fine-grained information while region-level calculation may lead to higher computational cost. Fortunately, under few-shot settings, there are only a few instances in each category for training which guarantees the computational efficiency and the whole process of generating query-specific prototypes is totally non-parametric.\par
\subsection{Nonlinear Classifier}

Given a generated prototype $\hat{\bm q}$ for $\tilde{\bm q}$, we introduce a relation function $\mathcal{H}$ to obtain their relation map:
\begin{equation}
\begin{aligned}
&{\bm R}=\mathcal{H}(\hat{\bm q}, \tilde{\bm q}),\\&{\bm r}_i = g(\hat{\bm v}_i, {\bm u}_i), i=1, \cdots, T,
\end{aligned}
\end{equation}
where ${\bm R}$ represents the relation map of $\hat{\bm q}$ and $\tilde{\bm q}$ while ${\bm r}_i$ is the $i$-$th$ element of ${\bm R}$. $\mathcal{H}$ indicates the relation function and $g$ denotes the same similarity function mentioned in Section 3.6, which will be discussed in detail in Section 5.2. To obtain the similarity score $\alpha_n$, we feed the relation map into the nonlinear classifier ${\mathcal{C}_{\Phi}}$, which is actually a four-layer MLP\cite{sung2018learning,hao2019collect} with learnable parameter ${\Phi}$:
\begin{equation}
\alpha_n={\mathcal{C}}_{\Phi}(\bm R).
\end{equation}\par

%algorithm
\begin{algorithm}[t]
\caption{Training of QPN} 
\label{T:QPN}
\begin{algorithmic}[1]
\Require
Episodic task $\mathcal{T}=\{ \mathcal{S}, \mathcal{Q}\}$;
\While {no converge}
\State $\vec{\mathcal{S}} \gets \mathcal{A}_{\Omega}[\mathcal{F}_{\Theta}(\mathcal{S})] $ 
\State $\vec{\mathcal{Q}} \gets \mathcal{A}_{\Omega}[\mathcal{F}_{\Theta}(\mathcal{Q})]$ 
\For {$\vec{\bm q}$ in $\vec{\mathcal{Q}}$ }
\For {$\vec{\bm s}$ in $\vec{\mathcal{S}}$}
\State Hierarchically represent $\vec{\bm s}$ and $\vec{\bm q}$ to $\tilde{\bm s}$ and $\tilde{\bm q} $ 
\State Generate QP $\hat{\bm q}$ for $\tilde{\bm q}$ from $\tilde{\bm s}$ by Eq. (6)
\State Get the relation map $\bm R$ by Eq. (7)
\State Obtain the relation score by Eq. (8)
\EndFor
\State Calculate probability $\mathcal{P}^{q}$ by softmax function
\EndFor 
\State $\mathcal{L} \gets -\sum \mathcal{Y}log(\mathcal P) $
\State Episodic Adam to minimize $\mathcal{L}$, update $\Theta, \Omega$ and $\Phi$
\EndWhile
\end{algorithmic}
\end{algorithm}
%algorithm
Note that for each query instance, there are $N$ similarity scores $[\alpha_1, \cdots, \alpha_N]$ representing the similarities between this query instance and $N$ support categories, respectively. \par
The whole training procedure of the proposed QPN is shown in Algorithm \ref{T:QPN}.

\section{Experiments}
%table
\begin{table*}[h]
\centering
\setlength{\tabcolsep}{2.3mm}{
\begin{tabular}{lcccccc}
\toprule
\multirow{2}{*}{\textbf{Model}} &\multirow{2}{*}{\textbf{Type}}& \multirow{2}{*}{\textbf{Embedding}}&\multicolumn{2}{c}{\textbf{\emph{mini}ImageNet}}&\multicolumn{2}{c}{\textbf{\emph{tiered}ImageNet}}\\
\cmidrule{4-7}
&&& \textbf{1-shot(\%)} & \textbf{ 5-shot(\%)}&\textbf{ 1-shot(\%)} & \textbf{ 5-shot(\%)} \\
\hline
\textbf{MAML}\cite{finn2017model-agnostic} &Optimization& Conv-32F & $48.70\pm1.84$ & $63.11\pm0.92$&$51.67\pm1.81$&$70.30\pm1.75$\\
%\textbf{MTL}$^\ast$\cite{sun2019meta}&Optimization&Conv-32F&$45.60\pm1.84$&$61.20\pm0.90$\\
% \textbf{Reptile}\cite{nichol2018on}& Conv-64F & $47.07\pm0.26 $& $ 62.74\pm0.37$\\
% \textbf{MAML++}$^\ast$\cite{antoniou2018train}&Meta&Conv-32F& $ 52.15\pm0.26$ & $68.32\pm 0.44 $\\
%\textbf{MM-Net}$^\ast$\cite{cai2018memory}& Meta& Conv-32F &$53.37\pm0.48$ &$66.97\pm0.35$\\
\textbf{MAML+L2F}\cite{baik2020learning} &Optimization&Conv-32F &$ 52.10\pm0.50$&$69.38\pm0.46 $&$54.40\pm0.50$&$73.34\pm0.44$\\ 
\textbf{MetaOptNet-RR}\cite{LeeMRS19}&Optimization&Conv-64F&${\color{blue}{53.23\pm0.59}}$&$69.51\pm0.48$&$54.63\pm0.67$&$72.11\pm0.59$\\
\hline
\textbf{MatchingNet}\cite{vinyals2016matching}&Metric&Conv-64F&$ 43.56\pm 0.84$ &$55.31\pm0.73$&-&-\\
\textbf{IMP}\cite{AllenSST19}&Metric&Conv-64F&$49.60\pm0.80$&$68.10\pm0.80$&-&-\\
\textbf{SAML}\cite{hao2019collect}&Metric&Conv-64F&$52.64\pm0.56$&$67.32\pm 0.75$&-&-\\
\textbf{DSN}\cite{simon2020adaptive}& Metric&Conv-64F& $51.78\pm0.96$&$68.99\pm0.69$&-&-\\
\textbf{GCR}\cite{li2019few-shot}&Metric& Conv-64F &$ 53.21\pm0.40$ & ${\color{blue}{72.34\pm0.32}}$&-&-\\
\textbf{TemperatureNet}\cite{0Temperature}&Metric&Conv-64F&52.39&67.89&-&-\\
\textbf{ProtoNet}\cite{snell2017prototypical} & Metric&Conv-64F &$ 49.42\pm0.78$ &$ 68.20\pm0.66$&$48.58\pm0.87$&$69.57\pm0.75$\\ 
%\textbf{GNN}$^\ast$\cite{garcia2017few-shot}& Metric&Conv-256F &$ 50.33\pm0.36$& $ 66.41\pm0.63$\\
\textbf{RelationNet}\cite{sung2018learning} & Metric&Conv-256F & $50.44\pm0.82$ & $65.32\pm0.70$&$54.48\pm0.93$&$71.31\pm0.78$\\
\textbf{CovaMNet}\cite{li2019distribution}&Metric&Conv-64F&$51.19\pm0.76$&$67.65\pm0.63$&${\color{blue}{54.98\pm0.90}}$&$71.51\pm0.75$\\
\textbf{DN4}\cite{li2019revisiting} & Metric&Conv-64F & $51.24 \pm0.74$& $71.02\pm0.64$&$53.37\pm0.86$&${\color{blue}{74.45\pm0.70}}$\\
%\textbf{ADM}$^\ast$\cite{LiWHSGL20}&Metric&Conv-64F&$54.26\pm0.63$ &$72.54\pm0.50$ \\
\hline
\textbf{QPN(Ours)($\xi$=5)} & Metric&Conv-64F&${\color{red}{54.92\pm0.82}}$&${\color{red}{73.18\pm0.81}}$&${\color{red}{55.96\pm0.77}}$&${\color{red}{75.01\pm0.84}}$\\
\toprule
\end{tabular}}
\caption{Comparisons with the state-of-the-art methods on \emph{mini}ImageNet and \emph{tiered}ImageNet with 95\% confidence intervals. The second column denotes the type of this method and the third column shows the type of the embeddings. Note that the best and the second best results are shown in red and blue respectively.}%Note that the results on \emph{mini}ImageNet are reported by the original works while the results on \emph{tiered}ImageNet are adopted from \cite{simon2020adaptive}.} Note that $^\ddagger$results need additional stage, e.g., pretraining and fine-tuning, etc, $^\dagger$results are re-implemented under the same setting and $^\ast$results are reported by original works. The best results are highlighted.}
\label{tabfs}
\end{table*}

\begin{table*}[h]
\centering
\setlength{\tabcolsep}{2.45mm}{
\begin{tabular}{lcccccc}
\toprule
\multirow{3}{*}{\textbf{Model}} & \multicolumn{6}{c}{\textbf{5-Way Accuracy($\%$)}}
\\
\cmidrule{2-7}
&\multicolumn{2}{c}{\emph{\textbf{Stanford Dogs}}} &\multicolumn{2}{c}{\emph{\textbf{Stanford Cars}}} 
&\multicolumn{2}{c}{\emph{\textbf{CUB-200-2011}}} \\
& 1-shot& 5-shot & 1-shot & 5-shot & 1-shot & 5-shot\\
\midrule
\textbf{MatchingNet}\cite{vinyals2016matching} & $35.80 \pm 0.99$ & $47.50 \pm 1.03 $ & $34.80 \pm 0.98$ & $44.70 \pm 1.03 $ & $61.16\pm 0.89$ & $72.86\pm 0.70$ \\
\textbf{ProtoNet}\cite{snell2017prototypical}& $37.59 \pm 1.00$ & $48.19 \pm 1.03 $ & $40.90 \pm 1.01$ & $52.93 \pm 1.03 $ & $51.31 \pm 0.91$ & $70.77 \pm 0.69$ \\
\textbf{MAML}\cite{finn2017model-agnostic} & $45.81 \pm 0.49$ & $60.01 \pm 0.28$ &$48.17 \pm 0.40$ & $61.85 \pm 0.26$ & $55.92\pm 0.95 $ & $ 72.09 \pm 0.76$ \\
\textbf{RelationNet}\cite{sung2018learning} & $44.49 \pm 0.39$ & $56.35 \pm 0.43$ &$48.59 \pm 0.45$ & $60.98 \pm 0.43$ & $62.45\pm 0.98 $ & $ 76.11 \pm 0.69$ \\
%\textbf{GNN}$^\dagger$\cite{garcia2017few-shot} & $46.98 \pm 0.98$ & $62.27 \pm 0.95 $ & $55.85 \pm 0.97$ & $71.25 \pm 0.89 $ & $51.83 \pm 0.98$ & $63.69 \pm 0.94$ \\
\textbf{CovaMNet}\cite{li2019distribution} & $\color{blue}{49.10 \pm 0.76} $ & $63.04 \pm 0.65 $ & $56.65 \pm 0.86$ & $71.33 \pm 0.62 $ & $60.58 \pm 0.69$ & $74.24 \pm 0.68$ \\
\textbf{DN4}\cite{li2019revisiting} & $45.41 \pm 0.76$ & $\color{blue}{63.51 \pm 0.62 }$ & $\color{blue}{59.84 \pm 0.80}$ & $\color{blue}{88.65 \pm 0.44} $ & $52.79 \pm 0.86$ & $\color{blue}{81.45 \pm 0.70}$ \\
\textbf{LRPABN}\cite{Huang2020Low}&$46.17\pm0.73$&$ 59.11\pm0.67$&$56.31\pm0.73$&$ 70.23\pm0.59$&$\color{blue}{63.63\pm0.77}$&$ 76.06\pm0.58$\\
\midrule 
\textbf{QPN(Ours)($\xi$=5)} & \color{red}{$53.69 \pm 0.62 $} & \color{red}{$70.98 \pm 0.70 $} & $\color{red}{63.91 \pm 0.58}$ &\color{red}{ $89.27 \pm 0.73$ } & $\color{red}{66.04 \pm 0.82}$ & \color{red}{ $82.85 \pm 0.76$ }\\
\toprule
\end{tabular}}
\caption{Comparisons with the state-of-the-art methods on three fine-grained datasets. Results are presented with 95\% confidence intervals. Note that the best and the second best results are shown in red and blue respectively.}
\label{tabfg}
\end{table*}
%In this section, we introduce the benchmark datasets and experimental settings of our proposed method. Then we make comparisons with recent effective approaches of both conventional few-shot classification and fine-grained few-shot classification. Finally, we present our ablation study and discussion on details.\par

\subsection{Datasets}
%In this paper, we evaluate our proposed method on five popular benchmark datasets.\par
\textbf{\emph{mini}ImageNet.} The \emph{mini}ImageNet dataset\cite{vinyals2016matching} is a subset of ImageNet\cite{deng2009imagenet:}, which consists of 100 classes and there are 600 instances with a resolution of $84\times 84$ in each class. Normally, we follow the dataset split procedure in \cite{li2019revisiting}, we divide the whole dataset into 64, 16 and 20 categories respectively for training, validation and test.\par
\textbf{\emph{tiered}ImageNet.} The \emph{tiered}ImageNet dataset\cite{RenTRSSTLZ18} is also a subset of ImageNet\cite{deng2009imagenet:} with 779, 165 samples and 608 categories. We split the dataset into 351, 97 and 160 categories for training, validation and test, respectively.\par

\textbf{\emph{Stanford Dogs}.} The \emph{Stanford Dogs} dataset\cite{khosla2011novel} was originally applied for fine-grained image classification, which consists of 120 categories and totally 20, 580 instances. Following the split of \cite{li2019revisiting}, we divide the dataset into 70 training categories, 20 validation categories and 30 test categories for fine-grained few-shot classification.\par
\textbf{\emph{Stanford Cars}.} The \emph{Stanford Cars} dataset\cite{krause20133d} was designed for fine-grained classification with 196 classes and 16, 185 images in all. Identically, we follow \cite{li2019revisiting} and split it into three parts(e.g., training, validation and test) with 130, 17 and 49 classes, respectively.\par
\textbf{\emph{CUB-200-2011}.} The \emph{CUB-200-2011} dataset\cite{wah2011caltech} contains 200 kinds of birds and totally 11, 788 instances. Following \cite{chen2019closer}, we split it into 100, 50 and 50 classes for training, validation and test.\par
%Note that all instances in the five datasets are rescaled to $84\times 84$ in resolution.\par

\subsection{Implementation Details}
\textbf{Network Architecture.} 
The network of QPN contains three parts: an embedding nerwork ${\mathcal{F}}_{\Theta}$, an Attention module ${\mathcal{A}}_{\Omega}$ and a nonlinear classifier ${\mathcal{C}}_{\Phi}$. For the fair comparisons with other methods, we choose commonly adopted four-layer CNN (Conv-4) with four convolutional blocks as our embedding network. Specifically, each convolutional block consists of a convolutional layer with 64 filters of size $3\times 3$, a batch normalization layer and a Leaky ReLU layer. Besides, we add $2\times 2$ max-pooling layer respectively after the middle two convolutional blocks. 
The architecture of the Attention module\footnote{In actual operation, we place the Attention module after the first convolutional block of embedding to capture more detailed information, which is experimentally effective.} ${\mathcal{A}}_{\Omega}$ consists of a two-layer MLP for the channel attention and a convolutional layer with the kernel size of $1\times 1$ for the spatial attention.
The nonlinear classifier ${\mathcal{C}}_{\Phi}$ is actually a four-layer MLP with the learnable parameter $\Phi$. Note that the only hyper-parameter in our model is the generation coefficient $\xi$, which is detailedly discussed in Section 5.1.\par
\textbf{Experimental Settings.}
Normally, we implement our experiments under the framework of Pytorch\cite{Paszke19}. Based on the episodic training mechanism, we train and test our model on a series of $C$-way $K$-shot tasks. During the training stage, we randomly sample 250, 000 episodes from the training set. For 5-way 1-shot tasks, there are 5 instances in the support set with 15 query instances per class. For 5-way 5-shot tasks, there are 25 instances in the support set with 15 query instances per class. Note that we choose the commonly adopted cross entropy loss and apply Adam\cite{kingma2015adam} as the optimizer during the training stage. We set the initial learning rate to 0.001 and halve it per 50, 000 episodes.\par
During the test stage, we randomly sample 600 episodes from the test set to evaluate the classification performance of our QPN and repeat this procedure five times. The final mean accuracy is reported with 95\% confidence intervals. The whole model of our QPN is trained in an end-to-end manner without additional stage.\par
%\subsection{Baseline Methods}

%To evaluate our proposed CPR-Net on the \emph{mini}ImageNet, we make comparisons with fifteen state-of-the-art models, including meta-learning based nethods: MAML\cite{finn2017model-agnostic}, MAML++\cite{antoniou2018train}, MTL\cite{sun2019meta}, MetaOpt\cite{LeeMRS19}, MM-Net\cite{cai2018memory}, and MAML+L2F\cite{baik2020learning}, and metric-learning based methods: Matching Net\cite{vinyals2016matching}, Prototypical Net\cite{snell2017prototypical}, GNN\cite{garcia2017few-shot}, Relation Net\cite{sung2018learning}, CovaMNet\cite{li2019distribution}, DN4\cite{li2019revisiting}, SAML\cite{hao2019collect}, DSN\cite{simon2020adaptive} and GCR\cite{li2019few-shot}.\par
%, ATL\cite{2020Learning}

%As for the fine-grained few-shot classification, we compare our CPR-Net with six few-shot methods on three fine-grained datasets, including Matching Net\cite{vinyals2016matching}, MAML\cite{finn2017model-agnostic}, Prototypical Net\cite{snell2017prototypical}, Relation Net\cite{sung2018learning}, CovaMNet\cite{li2019distribution} and DN4\cite{li2019revisiting}.
%and ATL\cite{2020Learning}.
\subsection{Comparisons with the SOTA Methods}
\textbf{Few-Shot Image Classification on \emph{mini}ImageNet and \emph{tiered}ImageNet.}
As is shown in Table \ref{tabfs}, we make comparisons with both optimization-based and metric-based methods on \emph{mini}ImageNet and \emph{tieredImageNet}. \par
For \emph{mini}ImageNet, under both 1-shot and 5-shot settings, our method achieves 54.92\% with 1.59\% improvement and 73.18\% with 0.84\% improvement from the second best methods, respectively. Notice that our QPN perform much better than the dynamic prototype based methods: IMP\cite{AllenSST19} and TemperatureNet\cite{0Temperature}. For \emph{tiered}ImageNet, our method also improves the classification performance by 0.98\% and 0.56\% under both settings.
The great improvements indicate the superiority of our method that generates high-quality region-level query-specific prototypes by searching out the local semantic consistency between each query sample and support categories.\par
\textbf{Fine-Grained Few-Shot Image Classification.}
As is shown in the Table \ref{tabfg}, we compare our QPN with seven FSL methods on three fine-grained datasets. We can see that our method outperforms all the methods presented under both 1-shot and 5-shot settings. In detail, on \emph{Stanford Dogs} dataset, our method is 4.59\% and 7.47\% better than the second best results under 1-shot and 5-shot settings; on \emph{Stanford Cars} dataset, QPN gains 4.07\% and 0.62\% improvements over the second best methods under both settings; on \emph{CUB-200-2011} dataset, our proposed method precedes the second best method with 2.41\% and 1.40\% promotion. These great improvements reveal the advantages of our proposed QPN on fine-grained few-shot image classification tasks. \par
Our QPN utilize the region-level semantic relevance to generate query-specific prototypes and achieves local semantic alignment between each query sample and support samples, which is naturally suitable for fine-grained image classification tasks. Besides, the attention mechanism also contributes to the performance on fine-grained tasks by highlighting discriminative key regions.%Our method adopts pixel-level consistency searching between query instances and support classes and conducts pixel-level semantic alignment, which naturally pay more attention to discriminative fine-grained information for classification. Moreover, our CPR-Net also efficiently filters out interferences that may cause misguidance to classifier by Attention Module.\par
\subsection{Ablation Study}

To further prove the effectiveness of our method, we conduct an ablation study on \emph{mini}ImageNet (see in Table \ref{tab3}).
Specifically, to show the impact of each component, we compare our entire QPN with its several incomplete versions. Our baseline model is similar to RelationNet\cite{sung2018learning}, which consists of an embedding network (Conv-4) and a four-layer MLP worked as a nonlinear classifier. We evaluate our three main contributions, i.e., Attention mechanism $\bm{\mathcal{A}}_{\Omega}$, Hierarchical Representation mechanism (denoted as \bm{$Hr$}) and the Prototype Generator (denoted as \bm{$Pg$}).\par %Note that $\mathcal{D}_{c}$ cannot be independent from $\mathcal{C}_{pr}$.\par
%Note that we use class mean prototypes to represent support classes and adopt the same relation function $\mathcal{H}$ in Section 3.6 to obtain the relation maps of the query instances and the mean prototypes. 
%table
\begin{table}[t]
\centering

\setlength{\tabcolsep}{3.6mm}{

\begin{tabular}{lcccc}
\toprule
\multicolumn{3}{l}{ \textbf{Ablation Models} \bm{$(\xi = 5)$}} &\multicolumn{2}{c}{\textbf{5-way Accuracy (\%)}} \\
\hline
$\bm{{\mathcal{A}}_{\Omega}}$ &\bm{$Hr$}&\bm{$Pg$}& \textbf{1-shot} &\textbf{5-shot}\\
\hline
& & & 50.04 & 68.59 \\
& \bm{$\surd$}& & 50.42 &68.96 \\
\bm{$\surd$}& & & 51.36 & 69.53 \\
&& \bm{$\surd$}& 52.36 & 70.83 \\
\bm{$\surd$}&\bm{$\surd$}& & 51.59 &69.65 \\
&\bm{$\surd$}&\bm{$\surd$}&53.46 &71.81\\
\bm{$\surd$}&&\bm{$\surd$}&53.99&72.35\\
\hline
\bm{$\surd$}&\bm{$\surd$}&\bm{$\surd$} & \color{red}{$54.92 $} &\color{red}{$73.18 $}\\
\toprule
\end{tabular}}

\caption{The results of the ablation study on \emph{mini}ImageNet under 1-shot and 5-shot settings. The best result is shown in red.}
\label{tab3}
\end{table}
%table
The results of the ablation study shown in Table \ref{tab3} definitely indicate the indispensability of the three components. %In detail, if we only remove ${\mathcal{A}}_{\Omega}$ from CPR-Net, the accuracies will decline by 1.46\% and 1.37\% under both settings. This proves the effectiveness of the ${\mathcal{A}}_{\Omega}$, which highlights key regions and compresses useless background information. Moreover, if we remove $\mathcal{D}_{c}$ from CPR-Net, the performance will suffer 0.93\% and 0.83\% decline on 1-shot and 5-shot tasks, which indicates that dimensional mismatching is alleviated by $\mathcal{D}_{c}$. On the basis of removing $\mathcal{D}_{c}$, we further remove $\mathcal{C}_{pr}$ and the results will suffer 2.63\% and 1.82\% decline under 1-shot and 5-shot settings, respectively. The gap reveals the effectiveness of $\mathcal{C}_{pr}$, which overcomes the biases between query instances and achieves positional semantic alignment at the same time.
\section{Further Analysis} 
\subsection{Influence of the Generation Coefficient \bm{$\xi$}}
\begin{figure}[t]
\centering
\includegraphics[width= 0.48\textwidth]{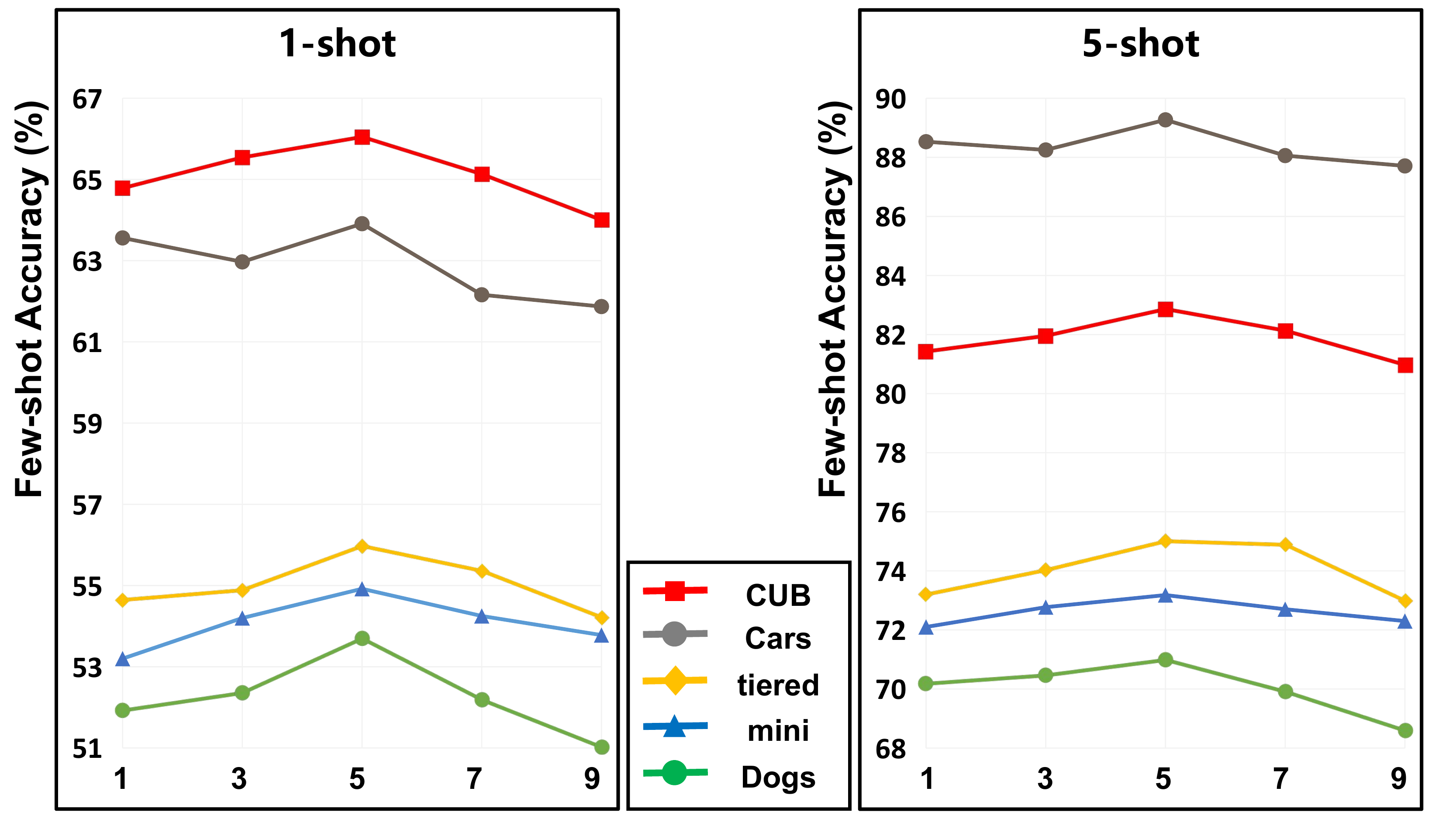}\\
\caption{Influence of the generation coefficient ($\xi=\{1, 3, 5, 7, 9\}$) on five datasets under both 1-shot and 5-shot settings.}
\label{fig4}
\end{figure}

In the Prototype Generator, we need to search out the $\xi$ most similar FGCs in one support class for each FGC of the query instance and generate a query-specific prototype. Since the PG is non-parametric, as the only hyper-parameter, the value of the generation coefficient $\xi$ becomes particularly significant, which directly affects the quality of the generation. To find out the relationship between $\xi$ and the final performance, we conduct a contrast experiment on five datasets under both 5-way 1-shot and 5-way 5-shot settings by changing the value of $\xi$ from 1 to 9. Experimental results are shown in Figure \ref{fig4}. It can be seen that the value of $\xi$ has a moderate influence on classification performance and we should choose a proper $\xi$ for each task.\par
%figure5

\subsection{Influence of the Similarity Function}
In Section 3.5 and Section 3.6, we use the similarity function $g$ to calculate the similarities between FGCs, so the choice of the similarity function can influence the final performance of our method to some extent. The similarity function has many choices, we choose two commonly adopted metrics\cite{hao2019collect} for comparison:\\
\textbf{Gaussian similarity}: \\
\begin{equation}
g(a,b) = e^{{a}\cdot{b}}.
\end{equation}
\textbf{Cosine similarity}:\\
\begin{equation}
g(a,b) = \frac{{a}^{\top} b}{||a||\cdot ||b||}.
\end{equation}
The results with different similarity functions are shown in Table \ref{metric}, it can be seen that the common similarity functions can perform well and we adopt the better cosine similarity in our work.\par
\subsection{Number of the Trainable Parameters}
To further verify the efficiency of our QPN, we compare the number of trainable parameters with other methods (see in Table \ref{tab4}). ProtoNet\cite{snell2017prototypical} and DN4\cite{li2019revisiting} have the smallest number of parameters due to their simple network architectures. RelationNet\cite{sung2018learning} introduces additional parameters in CNN based nonlinear classifier to boost the performance. GCR\cite{li2019few-shot} adopts the additional architectures with large amount of parameters. In contrast, QPN introduces a relatively smaller number of trainable parameters while achieves better performance. Compared with the Inception Operator, our proposed hierarchical representation strategy achieves the competitive performance.\par
\subsection{Cross-Domain FSL}
Cross-domain problems can better evaluate the ability to generalize to novel tasks of the model. Following the instructions of \cite{chen2019a}, we conduct a cross-domain experiment by training on \emph{mini}ImageNet and testing on \emph{CUB-200-2011}.\par The experimental results in Table \ref{Cdfsl} show that our QPN achieves better generalization performance across domains than the other approaches and even beats the baseline method with a stronger backbone. This shows the superiority of our method, which is good at learning transferable knowledge. For one thing, our QPN adopts the CBAM to highlight key regions and filter out the background information, which effectively filters the noise between different domains; for another, the PG generates targeted features for discriminative dominant regions, which can maintain an efficient metric when facing the serious domain shift.
%\subsection{Time Complexity}
%In the process of searching out the pixel-level semantic consistency, we have to calculate the similarities of all LRV pairs from support classes and query instances and conduct sort operation. In general, this will bring about amounts of computations. Fortunately, under few-shot settings, the proposed CPR-Net can exploit the pixel-level relevance without too much consideration about computational complexity. Specifically, during the training stage on \emph{mini}ImageNet, the time cost of our proposed CPR-Net under 5-way 1-shot/5-way 5-shot setting is 7ms or 13ms per episode with both 75 query instances on an Nvidia RTX 2080Ti GPU.\par

%table metric
\begin{table}[t]
\centering
\setlength{\tabcolsep}{5.2mm}{
\begin{tabular}{lcc}
\toprule
\textbf{Metric} & \textbf{5-way 1-shot}& \textbf{ 5-way 5-shot} \\ 
\hline
\textbf{Gaussian}& $54.08\pm 0.78$ & $72.63\pm 0.72$\\
\textbf{Cosine} & $ \color{red}{54.92\pm0.82 }$ &$ \color{red}{73.18\pm 0.81} $ \\
\toprule
\end{tabular}}
\caption{The performance with different similarity functions on \emph{mini}ImageNet.}
\label{metric}
\end{table}
%table para
\begin{table}[t]
\centering
\setlength{\tabcolsep}{1.4mm}{
\begin{tabular}{lccc}
\toprule
\textbf{Model} & \textbf{Params}& \textbf{ 1-shot}&\textbf{5-shot} \\ 
\hline
\textbf{ProtoNet}\cite{snell2017prototypical} & \color{red}{0.113M} & $49.42\pm0.78$&$ 68.20\pm0.66$\\
\textbf{DN4}\cite{li2019revisiting}&\color{red}{0.113M}& $51.24 \pm0.74$&$71.02\pm0.64$\\
\textbf{RelationNet}\cite{sung2018learning} & 0.229M & $50.44\pm0.82$ & $65.32\pm0.70$\\
%\textbf{GNN}\cite{garcia2017few-shot} & 1.619M & $ 50.33\pm0.36$ \\
\textbf{GCR}\cite{li2019few-shot}& 1.755M & \color{blue}{$53.21\pm0.40$} &{$72.34\pm0.32$}\\
%\textbf{ADM}\cite{LiWHSGL20}&\textbf{0.113M} & $54.26\pm0.63$ \\
\hline
\textbf{QPN (Ours)} & \color{blue}{ 0.150M} & $\color{red}{54.92\pm0.82}$&\color{blue}{$73.18\pm0.81 $} \\
\textbf{QPN}$_{io}$ \textbf{(Ours)}& 0.351M&$54.73\pm 0.75$&\color{red}{$73.36\pm 0.76$}\\
\toprule
\end{tabular}}
\caption{The number of the trainable parameters in different models and their performance on \emph{mini}ImageNet. Note that in QPN$_{io}$ we replace the hierarchical representation with the Inception Operator\cite{2015Going}.}
\label{tab4}
\end{table}

%table metric
\begin{table}[t]
\centering
\setlength{\tabcolsep}{2.5mm}{
\begin{tabular}{lcc}
\toprule
%\multicolumn{3}{c}{\textbf{\emph{mini}ImageNet$\to $\emph{CUB}}}\\
\textbf{Model} &\textbf{Embedding} & \textbf{5-way 5-shot(\%)} \\ 
\hline
\textbf{Baseline} \cite{chen2019a}& Resnet-18& \color{blue}{$65.57\pm0.70$}\\
\textbf{Baseline++} \cite{chen2019a} & Resnet-18 &$62.04\pm0.76$ \\
\hline
\textbf{MatchingNet} \cite{vinyals2016matching} & Resnet-18 &$53.07\pm0.74$ \\
\textbf{ProtoNet} \cite{snell2017prototypical} & Resnet-18&$62.02\pm0.70$ \\
\textbf{MAML} \cite{finn2017model-agnostic} & Resnet-18 &$51.34\pm0.72$ \\
\textbf{RelationNet} \cite{sung2018learning} & Resnet-18 &$57.71\pm0.73$ \\
\hline
\textbf{CovaMNet} \cite{li2019distribution} & Conv-64F &$63.21\pm0.68$ \\
\textbf{DN4} \cite{li2019revisiting} & Conv-64F &$63.42\pm0.70$ \\
\textbf{QPN (Ours)} & Conv-64F &\color{red}{$66.19\pm0.86$} \\
\toprule
\end{tabular}}
\caption{The performance of the cross-domain experiments (\textbf{\emph{mini}ImageNet$\to$ \emph{CUB}}). We adopt the results of the Resnet-18 reported by \cite{chen2019a}.}
\label{Cdfsl}
\end{table}

\section{Conclusion}
We propose a \emph{Hierarchical Representation based Query-Specific Prototypical Network} (QPN) for few-shot image classification, which generates query-specific prototypes by searching out the local semantic consistency between each query instance and support categories. Experimental results on both conventional and fine-grained datasets show the superiority of our model. The cross-domain experiment further ensures the stability of our method.

{\small
\bibliographystyle{ieee_fullname}
\bibliography{egbib}
}

\end{document}